%% file: main.tex
\documentclass{article} % For LaTeX2e
\usepackage[final]{colm2025_conference}

\usepackage{microtype}
\usepackage{hyperref}
\usepackage{url}
\usepackage{booktabs}

\usepackage{lineno}

\definecolor{darkblue}{rgb}{0, 0, 0.5}
\hypersetup{colorlinks=true, citecolor=darkblue, linkcolor=darkblue, urlcolor=darkblue}

\usepackage[T1]{fontenc}

\usepackage{graphicx}

\usepackage{array}
\usepackage{multirow}
\usepackage{booktabs} % For professional-looking tables
\usepackage{amsmath}
\usepackage{amssymb}
\usepackage{xcolor} % Required for text color

\title{Rethinking Agentic Workflows: Evaluating Inference-Based Test-Time Scaling Strategies in Text2SQL Tasks}

\author{
Jiajing Guo, Kenil Patel
\thanks{This work is done during internship at Bosch.}
, Jorge Piazentin Ono, Wenbin He, Liu Ren
\\
Bosch Research North America, USA \\
Bosch Center for Artificial Intelligence (BCAI)\\
\texttt{jiajing.guo@us.bosch.com} \\
\texttt{kenilpatel2047@gmail.com} \\
\texttt{\{jorge.piazentinono,wenbin.he2,liu.ren\}@us.bosch.com} \\
}

\begin{document}

\ifcolmsubmission
\linenumbers
\fi

\maketitle

\begin{abstract}
Large language models (LLMs) are increasingly powering Text-to-SQL (Text2SQL) systems, enabling non-expert users to query industrial databases using natural language. While test-time scaling strategies have shown promise in LLM-based solutions, their effectiveness in real-world applications, especially with the latest reasoning models, remains uncertain.
In this work, we benchmark \textbf{six} lightweight, industry-oriented test-time scaling strategies and \textbf{four} LLMs, including two reasoning models, evaluating their performance on the BIRD Mini-Dev benchmark. 
Beyond standard accuracy metrics, we also report inference latency and token consumption, providing insights relevant for practical system deployment.
Our findings reveal that Divide-and-Conquer prompting and few-shot demonstrations consistently enhance performance for both general-purpose and reasoning-focused LLMs. However, introducing additional workflow steps yields mixed results, and base model selection plays a critical role. This work sheds light on the practical trade-offs between accuracy, efficiency, and complexity when deploying Text2SQL systems.

\end{abstract}

\input{content/01.intro}

\input{content/03.results}
\input{content/05.discussion}

\bibliographystyle{colm2025_conference}
\bibliography{bib/paperpile}

\appendix
\section{Appendix}
\input{content/appendix.01}
\input{content/appendix.02}

\end{document}

%% file: content/01.intro.tex
\section{Introduction}
% Managing large-scale industrial data is a critical challenge in sectors such as manufacturing and energy, where actionable insights drive efficiency and innovation. 
% Text2SQL systems, particularly those powered by large language models (LLMs), have emerged to address this gap by translating natural language into SQL queries, offering adaptability and explainability without extensive retraining. 

The Text-to-SQL (Text2SQL) task aims to translate natural language questions into executable SQL queries. This capability is valuable in real-world scenarios, as it enables non-technical users to access and analyze information stored in enterprise databases without writing SQL.
Recent advances in large language models (LLMs) for instruction following and code generation have motivated researchers to employ \textbf{agentic methods} that leverage pretrained LLMs at inference time.
These approaches enhance model performance by applying a variety of test-time scaling strategies that go beyond basic Chain-of-Thought (CoT) prompting~\citep{Wei2022-CoT}, such as structured reasoning steps~\citep{Wang2023-MAC-SQL}, parallel execution~\citep{Lee2024-MCS-SQL}, verification~\citep{Xia2024-R3}, and result aggregation~\citep{Lee2024-MCS-SQL}. 
Instead of pure prompting, these methods scale the \textbf{reasoning process} by adopting a sequential workflow that decomposes query generation into modular steps like schema linking, SQL generation, and refinement.

% In real-world deployments, practitioners must consider factors beyond accuracy, including latency, stability, and token consumption. 
% Users expect Text2SQL systems to be both fast and reliable, and research shows that latency significantly affects user satisfaction. Although more complex workflows can improve accuracy, they often increase latency, a trade-off that is frequently overlooked in academic work.

Recent reasoning models, such as OpenAI’s o-series and Gemini 2.5, have demonstrated strong performance in logic and coding tasks. 
Unlike classical methods that rely on human-crafted reasoning frameworks, these models gain reasoning abilities through post-training techniques such as reinforcement learning or finetuning with reasoning procedures. 
This evolution raises key questions for practical Text2SQL deployment:
Is extensive prompt and workflow engineering still valuable given the advances in reasoning models?
Which test-time scaling strategies best balance accuracy and latency?
How should workflows be optimized for industry applications?

In this work, we focus on lightweight, industry-ready workflows rather than directly adopting the often complex workflows proposed in the literature. 
We evaluate \textbf{six} agentic workflows, each leveraging different test-time scaling strategies, across \textbf{four} LLMs, including both general and reasoning-focused models. 
Our assessment uses metrics such as SQL accuracy, latency, and token consumption, with the goal of providing actionable guidance for practitioners deploying Text2SQL systems.

Our main findings are:
\textbf{(1)} \textbf{Divide-and-Conquer (DC)} instructions and \textbf{few-shot demonstrations} significantly improve SQL generation quality. This holds true for both general-purpose models and models finetuned for advanced reasoning, indicating that even specialized models benefit from explicit, procedural guidance for Text-to-SQL tasks.
\textbf{(2)} Among more complex workflows, adding a \textbf{Result Verification} step provides the most consistent performance gains. However, increasing workflow complexity does not always lead to improvement in other workflows, and the effectiveness of any technique also depends on the base model.
\textbf{(3)} More complex workflows and challenging questions typically introduce higher latency, underscoring the need to balance efficiency and answer quality. In practice, this may require product design choices to manage user expectations.

%% file: content/03.results.tex
\section{Experiment}

Our experiments address the following research questions:

\textbf{RQ1}: How do inference-based test-time scaling strategies affect Text2SQL performance in general-purpose versus reasoning models?

\textbf{RQ2}: Which test-time scaling methods have the greatest impact on overall accuracy?

\textbf{RQ3}: What are the trade-offs between accuracy and system performance (e.g., latency) when applying these strategies?

\subsection{Agentic Workflows}
We selected six representative agentic workflows for evaluation. 
These workflows are extracted from literature on agent workflows for Text2SQL tasks and other reasoning and code tasks
\footnote{
We use the following symbols to represent workflow components:
\textbf{SW}: SQL Writer;
\textbf{EX}: Executor (a tool rather than an LLM agent);
\textbf{SR}: SQL Refiner (refines the SQL query based on executor output or agent feedback);
% \textbf{FP}: Feedback Provider;
% \textbf{KE}: Keyword Extractor;
% \textbf{ER}: Entity Retriever;
% \textbf{CR}: Column Retriever;
\textbf{>} : Sequential workflow order;
\textbf{<>}: Iteration between two components.
}
.
A summary of existing Text2SQL workflows can be found in Appendix~\ref{sec:text2sql_wf}, and diagrams of the workflows evaluated in our experiments are provided in Appendix~\ref{sec:wf_diagrams}.

\textbf{CoT + ReAct (Baseline): SW > EX <> SR}.
We choose ReAct agent~\citep{Yao2022-ReAct} as baseline workflow. 
Following the \textit{"think, act, observe"} loop, in Text2SQL system, agent iteratively refines the query if execution error or empty data is observed~\citep{Pourreza2023-DIN-SQL, Wang2023-MAC-SQL, Pourreza2024-CHASE-SQL, Xia2024-R3}.

\textbf{Divide-and-Conquer with and without Few-shots: SW > EX <> SR}.
Divide-and-Conquer is a reasoning technique that breaks down a complex problem into a series of smaller subproblems, solving them in sequence and combining them as the final response. 
This technique specifically instructs the agent how to perform decomposition and combination~\citep{Pourreza2024-CHASE-SQL}, and its effectiveness is evaluated with and without few-shot demonstrations.

\textbf{Parallel Scaling: 
(SW > EX <> SR) $\parallel$ 5 > MV / CS}. 
This workflow generated multiple candidates and select the final answer via majority vote. If the candidates have no majority, a \textbf{C}andidate \textbf{S}elector agent chooses the final answer.

\textbf{Result Verification:
SW > EX <> SR <> FP}.
This workflow aims to handle the situation where the generated SQL query is syntactically correct but semantically incorrect. After the workflow generates syntactically correct query, the execution output is passed to a \textbf{F}eedback \textbf{P}rovider to decide if it needs refinement.

\textbf{Retrieval-based Structured Reasoning: KE > (ER $\parallel$ CR) > SW > EX <> SR}.
Adapted from the CHESS method~\citep{Talaei2024-CHESS}, this workflow proceeds as follows.
First, a \textbf{K}eyword \textbf{E}xtractor identifies keywords in the question.
Next, two processes run in parallel: an \textbf{E}ntity \textbf{R}etriever uses the keywords to retrieve syntactically similar entities (database values) from a Locality-Sensitive Hashing (LSH) index, while a \textbf{C}olumn \textbf{R}etriever identifies semantically similar columns based on column descriptions.
The retrieved information is then passed to the \textbf{S}QL \textbf{W}riter, where selected tables and columns are emphasized by presenting \textit{Example Cell Values}. Non-selected tables and columns are represented only through their column descriptions in the database schema.

% The last three workflows are extensions of \texttt{DC 3-shot+ReAct} workflow and share its core setup, including the ReAct mechanism, prompt, and few-shot demonstrations. 

\subsection{Dataset and Metrics}

We evaluate on BIRD Mini-Dev benchmark~\citep{Li2023-BIRD}. 
% BIRD a large-scale cross-domain dataset consisting of 95 databases (totaling 33.4 GB) and 12,751 question–SQL pairs across 37 domains. 
BIRD Mini-Dev is a subset with 500 question-SQL pairs derived from the original BIRD Dev set.
Metrics include both accuracy and system performance, as described below.

% \textbf{Execution Accuracy (EX)}. The percentage of generated SQL queries that produce results identical to those of the ground truth on the target database~\cite{Li2023-BIRD}. This is one of the most commonly used metrics in Text2SQL tasks. We use it as the primary accuracy metric. Other performance metrics are in Appendix~\ref{sec:exp_details}.

\textbf{Soft F1-Score}. Evaluates SQL query correctness by measuring the similarity between the tables produced by the predicted queries and those generated by the ground-truth queries, thereby mitigating the impact of variations such as column order or missing values. This metric is less strict than the commonly used Execution Accuracy (EX).

\textbf{Execution Error Rate}. The percentage of tasks that encounter syntax execution error in the workflow. Since all workflows include a syntax-fixing iteration, this metric indicates the quality of generated SQL.

% \textbf{Soft F1-Score}. A less strict metric that assesses the accuracy of SQL query via measuring the similarity between the tables produced by predicted SQL queries and those generated by the ground truth SQL queries, mitigating the impact of variations such as column order and missing values.

% \textbf{Reward-based Valid Efficiency Score (R-VES)}. Quantifying how effectively models generate SQL queries that produce correct and optimized results. 
% % As a more stable and reliable version version of VES, reward point based on the time ratio in R-VES.  

\textbf{Inference Time}. The duration, in seconds, from the moment the agentic workflow receives the user's question to the generation of the corresponding SQL query
\footnote{Inference time can be affected by various factors, including model region, network latency, server resource availability, and the complexity of both the workflow and prompts. Therefore, the values reported in this paper should be interpreted as indicative rather than absolute, and may vary under different deployment conditions.}.

\textbf{Number of LLM Calls}. Average number of LLM calls used in the workflow. 

\textbf{Token Count}. The average number of prompt and completion tokens (in units of 1,000 tokens) required to generate a single SQL query.

\section{Results}

\input{content/fig.main_results}

% In this section, we report our experimental results for each research question. To gain a deeper understanding of model performance, we conducted an error analysis using the \texttt{o4-mini} model, categorizing failures based on common error types from previous work~\cite{Wang2023-MAC-SQL, Pourreza2023-DIN-SQL}.

\textbf{RQ1}: How do inference-based test-time scaling strategies affect Text2SQL performance in general-purpose versus reasoning models?

We observed that both reasoning and general-purpose models benefit from test-time scaling strategies, with the most significant performance gains stemming from combining Divide-and-Conquer (DC) and few-shot demonstration.
As shown in Figure~\ref{fig:soft_f1_ex_error_rate}, the application of \texttt{DC 3-shot+ReAct} workflow consistently enhances the Soft-F1 score for all models. 
For instance, the general-purpose model GPT-4o saw its Soft-F1 increase from 61.1 in the baseline to 64.4. Similarly, the reasoning model o4-mini improved from a baseline of 56.3 to 65.5. 
This indicates that although reasoning models have the capacity to scale their reasoning, incorporating human-like procedural steps for a specific task like SQL generation still improves their performance.

Another finding is that a strong foundational model can be more impactful than workflow complexity. 
For example, the Gemini 2.5 Flash model baseline (\texttt{CoT+ReAct}), with Soft-F1 of 65.75, already performs better than the most complex workflows of GPT-4o (max 64.95) and Gemini 1.5 Flash (max 63.63). 
% As Figure~\ref{fig:soft_f1_ex_error_rate} (right) shows, only Gemini 2.5 Flash showed a consistent decrease in errors as workflow complexity increased. In contrast, other models appeared to rely on trial and error to generate executable SQL queries, without a corresponding reduction in final errors. 
While workflow design is important for performance tuning, this result clearly indicates that the selection of a robust base model is a critical factor in achieving high-performance Text2SQL solutions.

\input{content/fig.error_analysis}

\textbf{RQ2}: Which test-time scaling methods have the greatest impact on overall accuracy?

We found that the combination of \textbf{Divide-and-Conquer}, \textbf{few-shot demonstrations} and \textbf{ReAct} (denoted as \textbf{DC 3-shot+ReAct}) consistently improves performance over the Baseline CoT+ReAct across all models.
Given its robust gains, we use \texttt{DC 3-shot+ReAct} as the primary benchmark to evaluate the impact of more complex workflows.
Adding additional steps to this strong baseline yielded mixed results.

The \textbf{Verification} method proved most effective, delivering reliable performance boosts across most models. 
It increased the Soft-F1 score for Gemini 1.5 Flash (from 62.58 to 63.63), Gemini 2.5 Flash (68.12 to 68.44), and GPT-4o (64.44 to 64.95). The only exception was o4-mini, where the score remained unchanged (65.53).
\textbf{Parallel scaling} showed moderate gains but slightly degraded performance for Gemini 2.5 Flash (67.32 vs. 68.12).
% showed moderate gains. The \texttt{DC 3-shotx5 Parallel} workflow also improved accuracy for three of the four models, including o4-mini, which achieved the highest score with this method (66.15). However, this approach slightly degraded performance for Gemini 2.5 Flash (67.32 vs. 68.12).
In contrast, the \textbf{retrieval-enhanced} method was generally counterproductive, underperforming \texttt{DC 3-shot+ReAct} workflow for nearly every model. 
This method is designed to help the model by filtering the database schema down to only the most relevant columns based on semantic similarity. However, this creates a critical trade-off: if identifying the correct columns requires complex reasoning rather than simple similarity matching, the model may be deprived of essential information, leading to errors.

To investigate this, we conducted an error analysis by prompting \texttt{o4-mini} model to categorize failures based on common error types summarized from previous work~\citep{Wang2023-MAC-SQL, Pourreza2023-DIN-SQL}.
Our analysis (Figure~\ref{fig:error_analysis}) reveals that Wrong Query Logic is the most common failure type across all methods and models, and the retrieval-enhanced method consistently exacerbates this problem. 
% For every model, Retrieval+DC 3-shot produced a higher proportion of logic errors than the DC 3-shot+ReAct workflow (e.g., increasing from 36.2\% to 41.0\% for Gemini 2.5 Flash). 
Furthermore, the retrieval method also increased the rate of Schema Linking Errors for Gemini 1.5 Flash (from 14.9\% to 16.4\%) and o4-mini (from 10.4\% to 12.2\%).

\textbf{RQ3}: What are the trade-offs between accuracy and system performance when applying these strategies?

Gemini Flash models demonstrate significantly lower latency, with response times between 5.02–12.03 seconds, compared to GPT-4o and o4-mini, which take 15.70–18.43 seconds (Figure~\ref{fig:latency_tokens}).
Analyzing the results by query difficulty reveals a clear trade-off. More challenging questions take longer to answer, consume more tokens, and often result in lower accuracy. 
A notable finding is that incorrect answers are 19.58\% slower to generate than correct ones. 
The extended wait times for these complex queries are particularly problematic, as they do not guarantee a correct answer.
The findings on latency and error rates highlight a critical challenge for practical deployment: balancing computational efficiency with output quality. Consequently, the focus may need to shift to product design to manage user expectations and experience. This could include allowing users to select between speed and accuracy modes or adding UX features such as progress indicators or partial results to manage expectations during longer waits.

%% file: content/fig.main_results.tex
\begin{figure}[!htbp]
  \centering
  % \begin{minipage}{0.49\textwidth}
  %   \centering
  %   \includegraphics[width=\linewidth]{content/figures/execution_accuracy.png}
  %   \caption{Execution Accuracy.}
  %   \label{fig:ex}
  % \end{minipage}
  \begin{minipage}{0.48\textwidth}
    \centering
    \includegraphics[width=\linewidth]{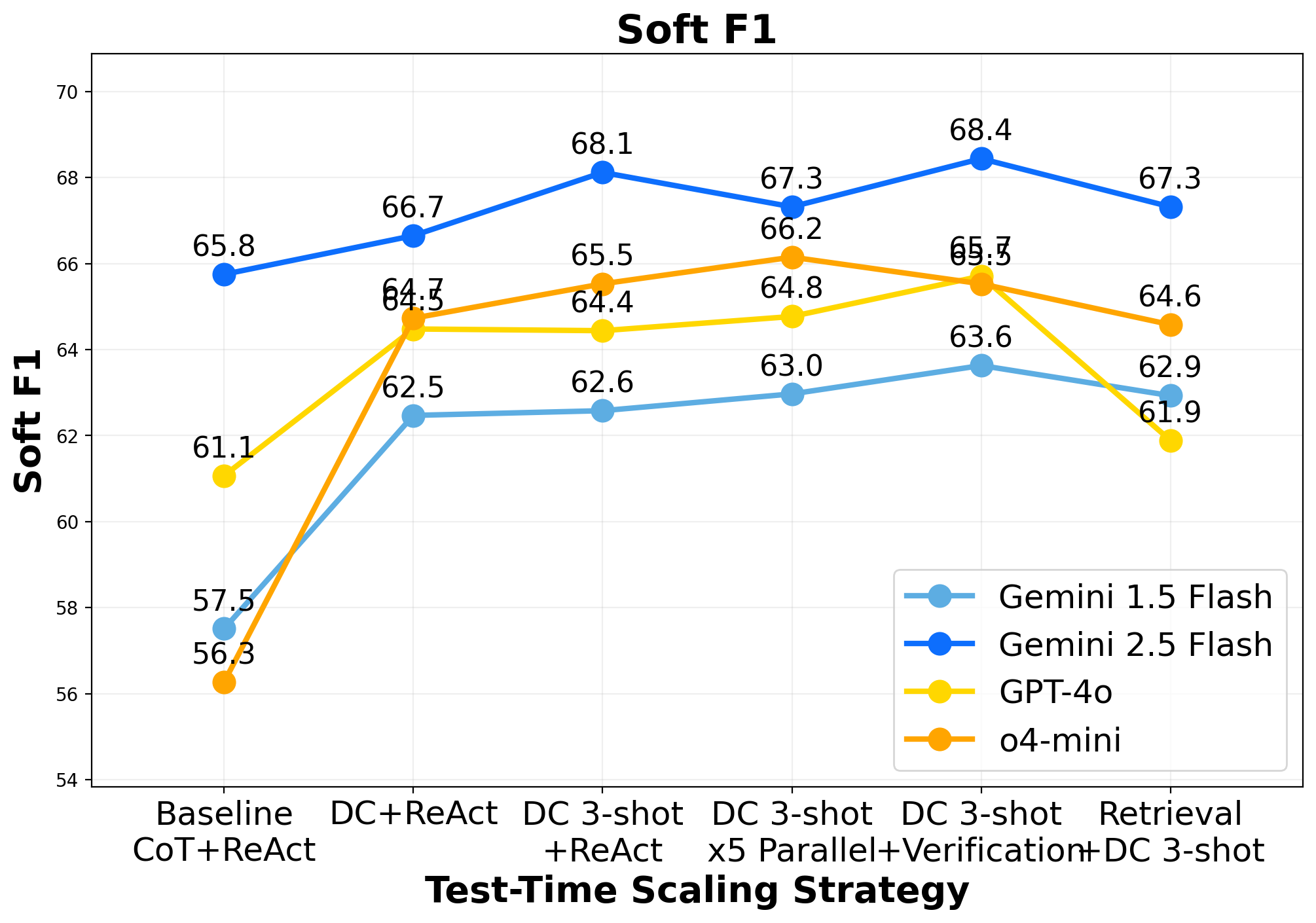}
    % \caption{Soft-F1 Score}
    % \label{fig:soft_f1}
    % \label{fig:soft_f1_ex_error_rate}
  \end{minipage}
  \hfill
  \begin{minipage}{0.48\textwidth}
    \centering
    \includegraphics[width=\linewidth]{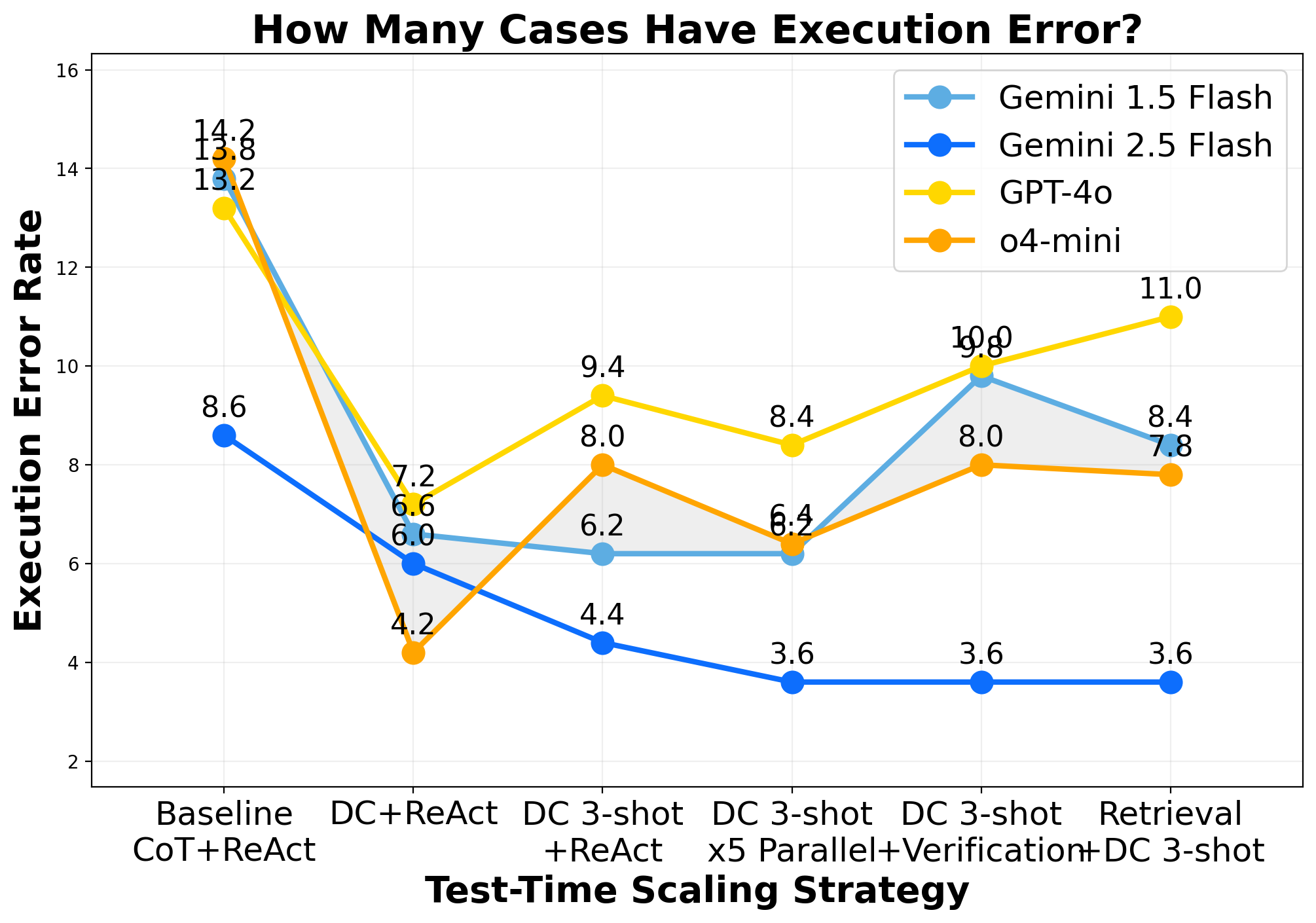}
    % \caption{Execution Error Rate.}
    % \label{fig:error_rate}
  \end{minipage}
\caption{Soft-F1 Score (left) and Execution Error Rate (right).
The label \texttt{DC 3-shot x5 Parallel}, \texttt{DC 3-shot+Verification}, and \texttt{Retrieval+DC 3-shot} denote the workflows that use parallel scaling, result verification, and retrieval-enhanced techniques, respectively. 
% All are extensions of \texttt{DC 3-shot+ReAct} workflow and share its core setup, including the ReAct mechanism, prompt, and few-shot demonstrations. 
For visual clarity, the "+ReAct" suffix is omitted from their labels in the figure.
}
\label{fig:soft_f1_ex_error_rate}
\end{figure}

\begin{figure}[!htbp]
  \centering
  \begin{minipage}{0.48\textwidth}
    \centering
    \includegraphics[width=\linewidth]{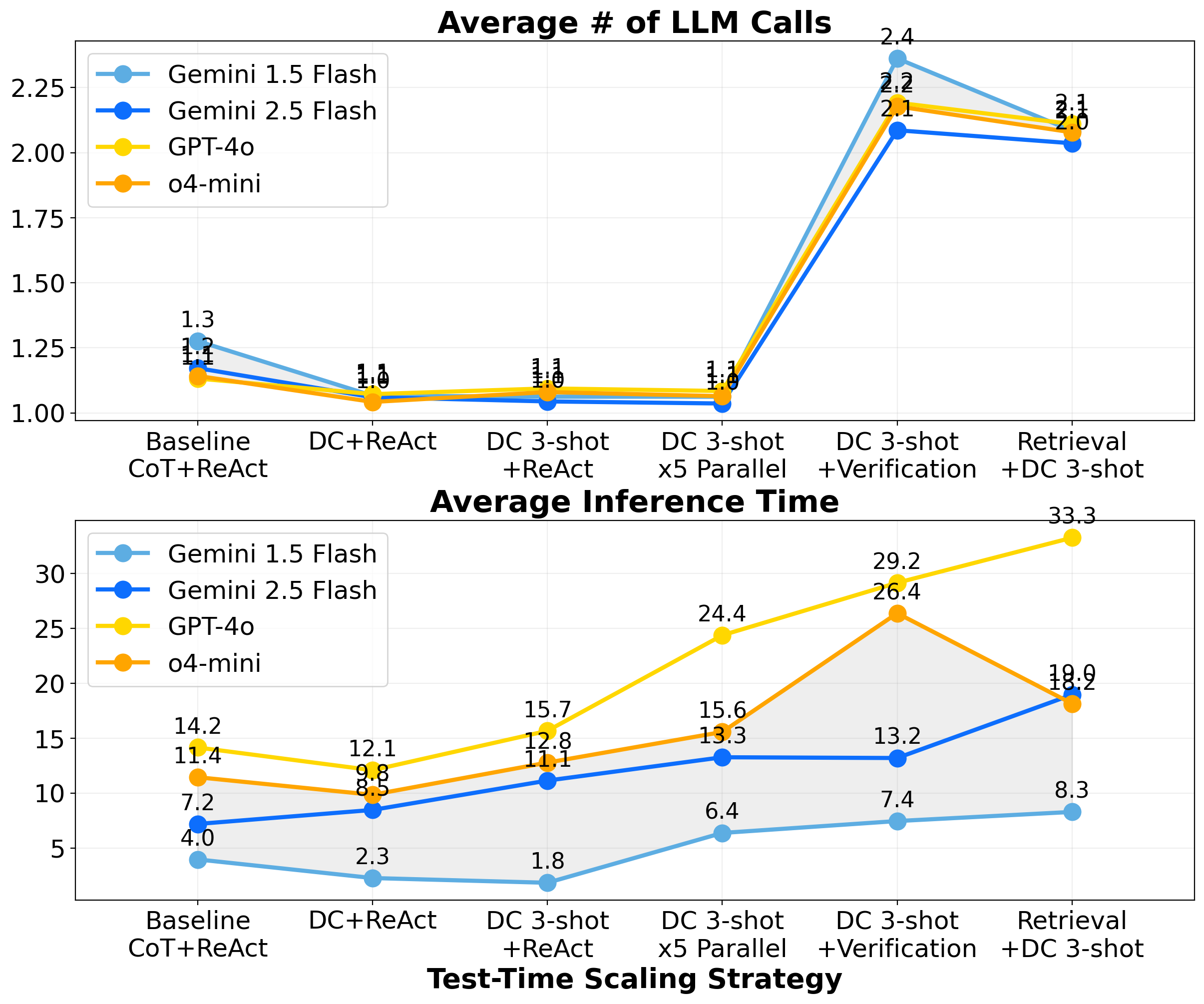}
  \end{minipage}
  \hfill
  \begin{minipage}{0.48\textwidth}
    \centering
    \includegraphics[width=\linewidth]{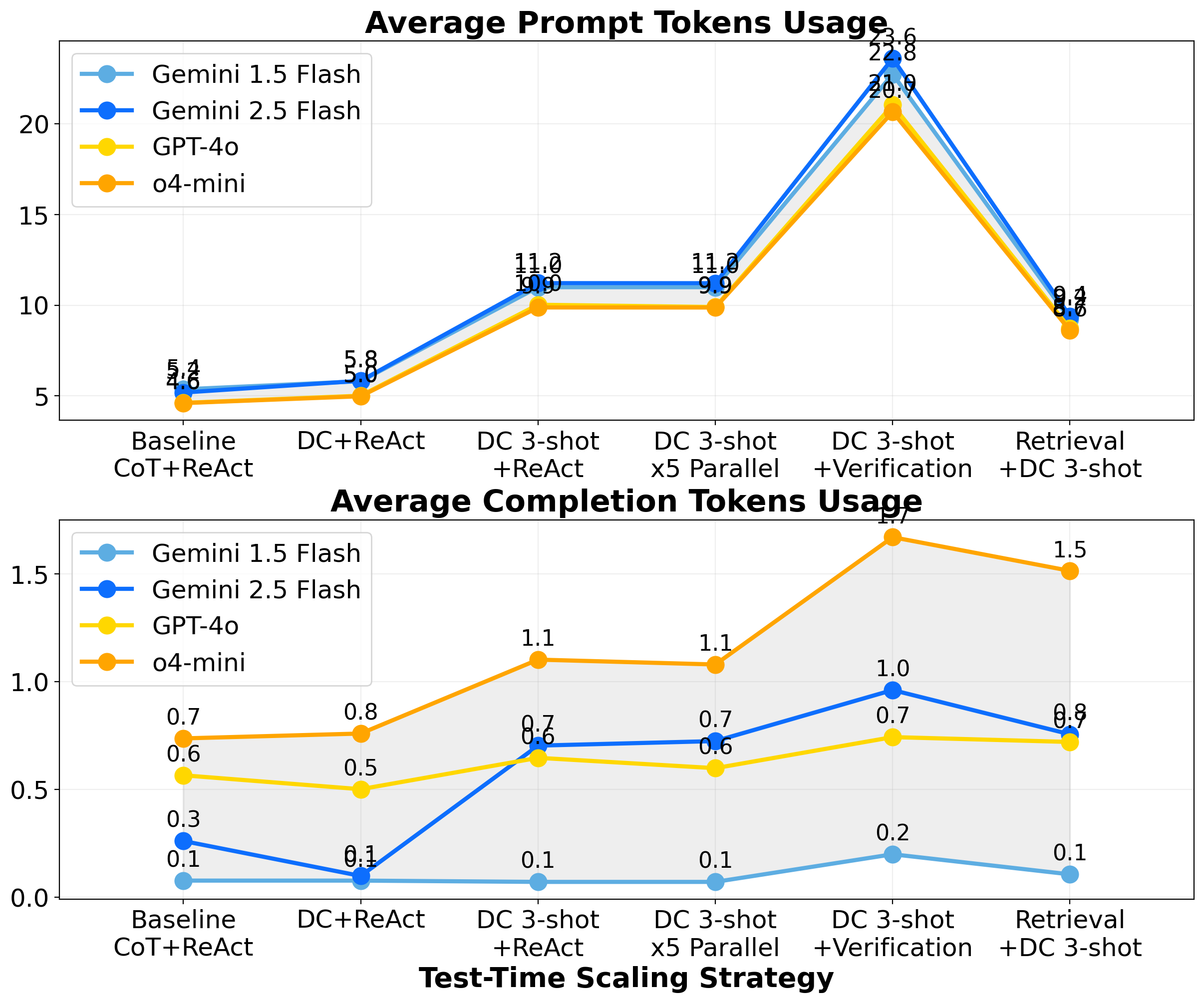}
  \end{minipage}
\caption{Average \# of LLM Calls and Latency (left). 
Prompt and Completion Tokens (right).}
\label{fig:latency_tokens}
\end{figure}

%% file: content/fig.error_analysis.tex
\begin{figure}
    \centering
    \includegraphics[width=0.98\linewidth]{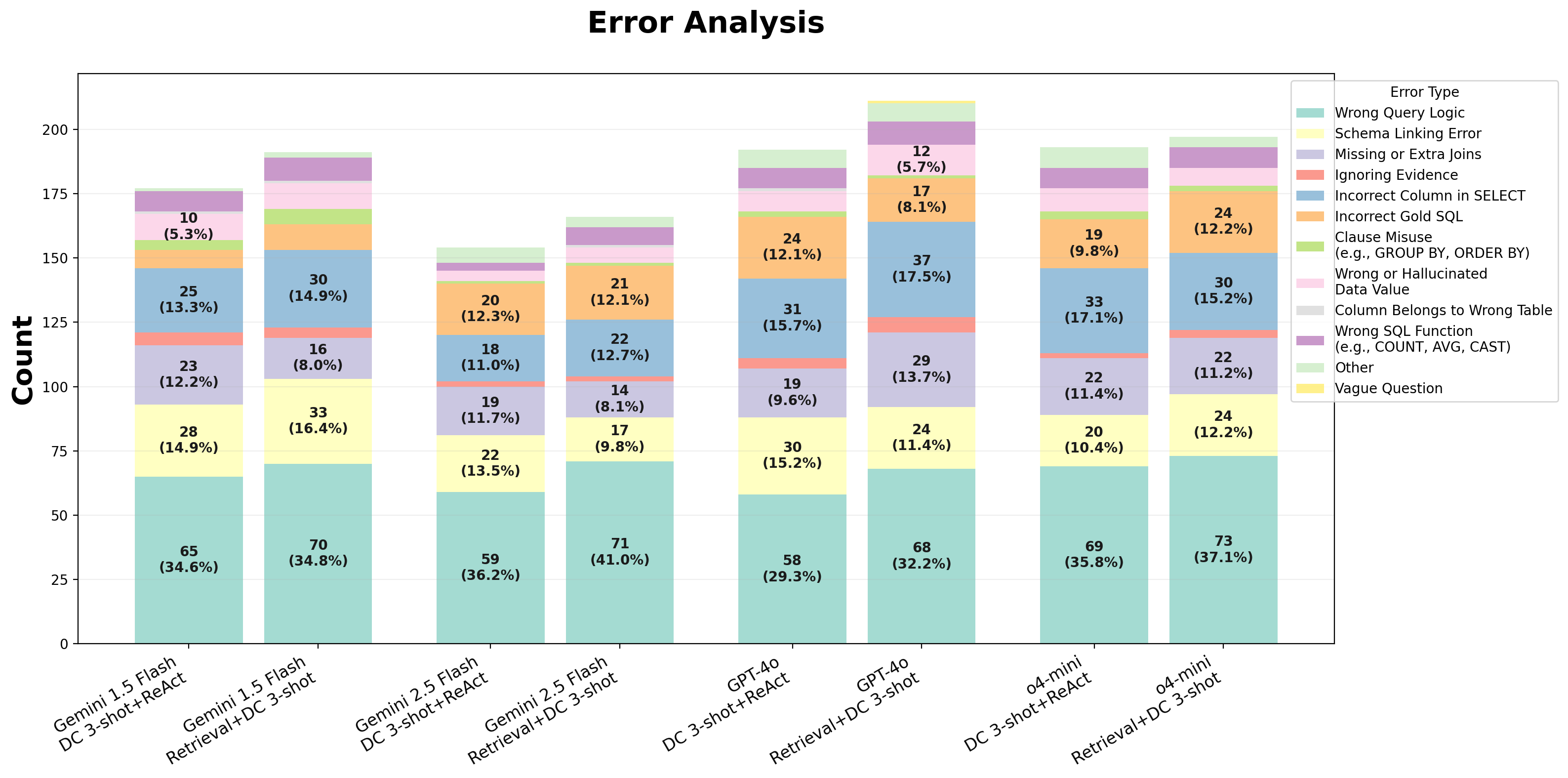}
    \caption{Error analysis on \textit{DC 3-shot+ReAct} and \textit{Retrieval+DC 3-shot+ReAct} workflows. 
    % Incorrect cases are defined following Execution Accuracy as it introduce a clear binary result.
    }
    \label{fig:error_analysis}
\end{figure}

% \begin{figure}[!htbp]
%   \centering
%   \begin{minipage}{0.49\textwidth}
%     \centering
%     \includegraphics[width=\linewidth]{content/figures/error_analysis_nl2sql-gpt-4o_stacked_bar.png}
%     \caption{Error analysis of GPT-4o}
%     \label{fig:latency}
%   \end{minipage}
%   \hfill
%   \begin{minipage}{0.49\textwidth}
%     \centering
%     \includegraphics[width=\linewidth]{content/figures/error_analysis_nl2sql-o4-mini_stacked_bar.png}
%     \caption{Error analysis of o4-mini}
%     \label{fig:tokens}
%   \end{minipage}
% \end{figure}

%% file: content/05.discussion.tex
\section{Conclusion}

% In this work, we evaluate six inference-based test-time scaling strategies on four LLMs for Text2SQL tasks. Advanced prompting and self-correct improve performance for both general LLMs and reasoning models, simple workflows that add upon baseline do not always help. 
% As a preliminary benchmark, this work has several limitations. We focused only on lightweight workflows, which may not capture the full potential of combined scaling strategies. 
% Evaluation was limited to the BIRD Mini-Dev for efficient testing; 
% Also, our reported latency values are approximate, as they are affected by factors like model hosting region and network conditions. 
% Future research should consider more complex workflows, diverse benchmarks, and other applications beyond Text2SQL.

In this work, we evaluate six inference-based test-time scaling strategies on four LLMs for Text2SQL tasks.
% This paper benchmarks the practical trade-offs between accuracy and efficiency for Text-to-SQL systems by evaluating six test-time scaling strategies on four different LLMs. 
We found that Divide-and-Conquer instructions and few-shot demonstrations, provide significant and reliable performance gains across all models, including those specialized for reasoning. This suggests that advanced models also benefit from explicit procedural guidance. In contrast, more complex workflows yielded mixed results: while a result verification step was beneficial, increasing complexity did not always lead to better outcomes. Also, a robust base model is a critical prerequisite for high performance.

While our findings offer actionable guidance for practitioners, this work serves as a preliminary benchmark with several limitations. We focused exclusively on lightweight, industry-oriented workflows, which may not represent the upper bounds of performance achievable with more sophisticated agentic workflow designs. Our evaluation was confined to the BIRD Mini-Dev benchmark for reasons of efficiency. Additionally, the reported latency and token consumption metrics should be interpreted as indicative rather than absolute, as they are influenced by external factors like network latency and server load. Future work can build on our results by examining more complex workflows, validating these findings across a broader range of datasets, and exploring the applicability of these strategies to other tasks.

%% file: content/appendix.01.tex
\subsection{Agentic Workflows in Text2SQL}
\label{sec:text2sql_wf}

We extract and generalize common agent modules (e.g., keyword extraction, column selection, SQL generation, self-repair) and workflow patterns such as sequential and parallel that have emerged as effective building blocks. 
Table~\ref{tab:workflow} and ~\ref{tab:workflow_symbol} summarize agentic workflows for NL2SQL tasks from previous literature. 

\input{content/table.workflows}

\subsection{Workflow Diagrams}
\label{sec:wf_diagrams}

\begin{figure}[htbp]
  \centering
  \begin{minipage}{0.3\textwidth}
    \centering
    \includegraphics[width=\linewidth]{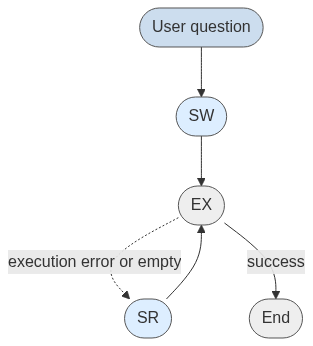}
    % \caption{SW > EX <> SR}
    % \label{fig:wf-sw_ex_sr}
  \end{minipage}
  \hfill
  \begin{minipage}{0.3\textwidth}
    \centering
    \includegraphics[width=\linewidth]{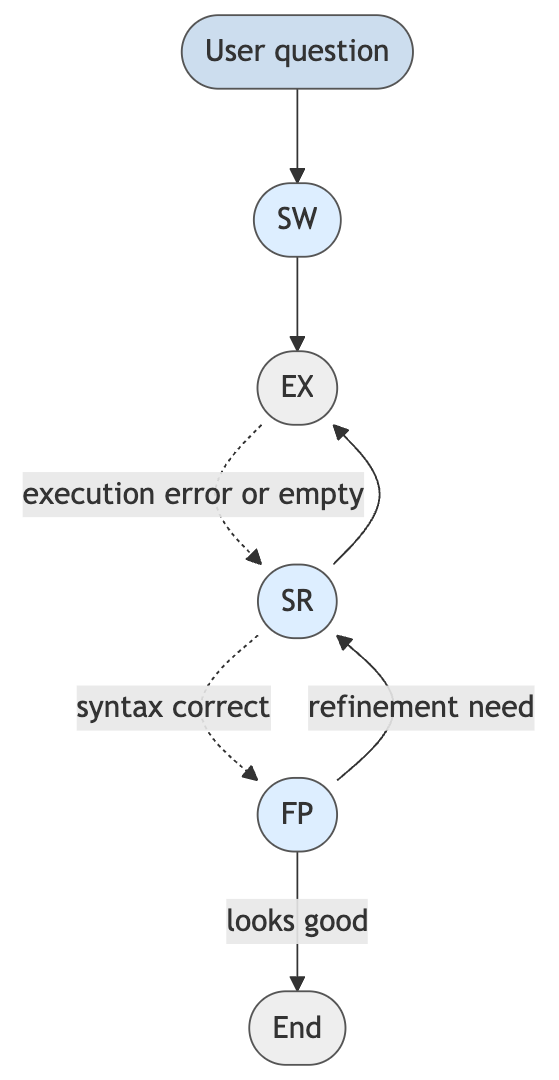}
    % \caption{SW > EX <> SR <> FP}
    % \label{fig:wf-sw_ex_sr_fp}
  \end{minipage}
  \begin{minipage}{0.3\textwidth}
    \centering
    \includegraphics[width=\linewidth]{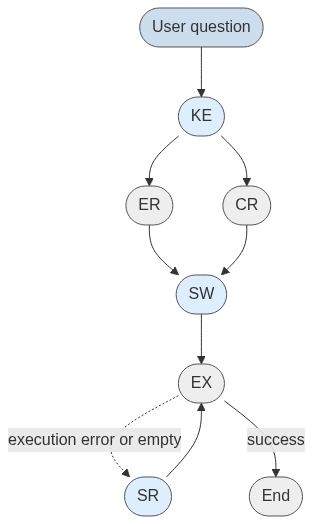}
    % \caption{KE > (ER $\parallel$ CR) > SW > EX <> SR}
    % \label{fig:wf-kwe_ercr_sw}
  \end{minipage}
    \caption{Workflows diagrams of 
    ReAct: SW > EX <> SR (left), 
    Verification: SW > EX <> SR <> FP (middle), 
    Retrieval-based: KE > (ER $\parallel$ CR) > SW > EX <> SR (right)
    .}
    \label{fig:workflows}
\end{figure}

Figure~\ref{fig:workflows} depicts the diagrams of three workflows in the study. The other workflow that uses parallel scaling technique runs five threads of the DC+3-shot. It selects the final answer via majority vote. If the candidates have no majority, a Candidate Selector agent chooses the final answer.

% \begin{figure}
%     \centering
%     \includegraphics[width=0.5\linewidth]{content/wf_figures/sw_ex_sr.png}
%     \caption{SW > EX <> SR}
%     \label{fig:wf-sw_ex_sr}
% \end{figure}

% \begin{figure}
%     \centering
%     \includegraphics[width=0.29\linewidth]{content/wf_figures/sw_ex_sr_fp.png}
%     \caption{SW > EX <> SR <> FP}
%     \label{fig:wf-sw_ex_sr}
% \end{figure}

% \begin{figure}
%     \centering
%     \includegraphics[width=0.29\linewidth]{content/wf_figures/sw_ex_sr_fp.png}
%     \caption{SW > EX <> SR <> FP}
%     \label{fig:wf-sw_ex_sr}
% \end{figure}

% \begin{figure}[htbp]
%   \centering
%   \begin{minipage}{0.49\textwidth}
%     \centering
%     \includegraphics[width=0.8\linewidth]{content/wf_figures/sw_ex_sr.png}
%     \caption{SW > EX <> SR}
%     \label{fig:wf-sw_ex_sr}
%   \end{minipage}
%   \hfill
%   \begin{minipage}{0.49\textwidth}
%     \centering
%     \includegraphics[width=0.8\linewidth]{content/wf_figures/sw_ex_sr_fp.png}
%     \caption{SW > EX <> SR <> FP}
%     \label{fig:wf-sw_ex_sr}
%   \end{minipage}
% \end{figure}

%% file: content/table.workflows.tex
\begin{table}[htbp]
  % Don't center the whole table, let minipages fill the width
  \begin{minipage}[t]{0.58\textwidth}
    % Remove \centering for left alignment
    \raggedright
    \resizebox{\textwidth}{!}{
    \begin{tabular}{@{}ll@{}}
      \toprule
      \textbf{Workflow} & \textbf{Reference} \\
      \midrule

      \textbf{CS} > \textbf{QC} > \textbf{SW} > \textbf{SR} 
        & DIN-SQL \\

      \textbf{CS} > \textbf{SW} > EX > \textbf{SR} 
        & MAC-SQL \\

      ES > \textbf{SW} 
        & DAIL-SQL \\

      (\textbf{SW} > \textbf{FR} > \textbf{SR}) $\parallel$ ...$_n$ 
        & R3 \\

      \textbf{CS} > \textbf{QC} > \textbf{SW} > \textbf{SR} > \textbf{FR} 
        & DEA-SQL \\

      (\textbf{SW} > EX) $\parallel$ ...$_n$ > \textbf{CP}
        & SQLPrompt \\

      \textbf{CS} > (\textbf{SW} $\parallel$ \textbf{SW} ) > \textbf{CP}
        & MCS-SQL \\

      \textbf{CS} > DC > (\textbf{SW} > \textbf{SR} $\parallel$ \textbf{SW} > \textbf{SR} $\parallel$ …\;) 
        & MAG-SQL \\

      \textbf{KE} > ER > \textbf{CS} > \textbf{SW} > EX > \textbf{SR}
        & CHESS \\

      \textbf{KE} > ER > (\textbf{SW} > EX > \textbf{SR} $\parallel$...)$_n$ > \textbf{CP}  
        & CHASE-SQL \\

      \textbf{CS} > \textbf{CE} > (\textbf{SW} $\parallel$ \textbf{SW}) > \textbf{SR} > \textbf{CP}  
        & ReFoRCE \\

      \bottomrule
    \end{tabular}
    }
    \caption{High-Level Workflows for LLM/Agent-Based Text-to-SQL Methods. Bold symbols indicate components powered by LLMs.}
    \label{tab:workflow}
  \end{minipage}%
  \hfill % this puts the tables to the extreme sides
  \begin{minipage}[t]{0.4\textwidth}
    \raggedleft
    \resizebox{\textwidth}{!}{
    \begin{tabular}{@{}ll@{}}
      \toprule
      \textbf{Symbol} & \textbf{Definition} \\
      \midrule
      KE  & Keyword Extractor \\
      CS  & Column Selector \\
      ES  & Example Selection \\
      ER  & Entity Retrieval \\
      QC  & Query Classifier / Decomposer \\
      DC  & Decomposer (task splitting) \\
      SW  & SQL Writer \\
      EX  & Executor (runs SQL) \\
      SR  & Self-Refiner (LLM error correction) \\
      FR  & Feedback Refiner \\
      CE  & Column Explorer \\
      CP  & Candidate Picker / Selector \\
      $\parallel$ & Indicates parallel execution \\
      ( )  & Groups parallel branches in workflow \\
      \bottomrule
    \end{tabular}
    }
    \caption{Standardized Symbol Definitions and Notation for Workflow Mapping.}
    \label{tab:workflow_symbol}
  \end{minipage}
\end{table}

%% file: content/appendix.02.tex
\subsection{Experiment Result Details }
\label{sec:exp_details}

Besides Soft-F1 score, we also report other commonly used metrics in Text2SQL tasks. 

% \textbf{Soft F1-Score}. A less strict metric that assesses the accuracy of SQL query via measuring the similarity between the tables produced by predicted SQL queries and those generated by the ground truth SQL queries, mitigating the impact of variations such as column order and missing values.

\textbf{Execution Accuracy (EX)}. The percentage of generated SQL queries that produce results identical to those of the ground truth on the target database~\cite{Li2023-BIRD}. This is one of the most commonly used metrics in Text2SQL tasks. We use it as the primary accuracy metric. Other performance metrics are in Appendix~\ref{sec:exp_details}.

\textbf{Reward-based Valid Efficiency Score (R-VES)}. Quantifying how effectively models generate SQL queries that produce correct and optimized results. 
As a more stable and reliable version version of VES, reward point based on the time ratio in R-VES.  

\begin{figure}[htbp]
  \centering
  % \begin{minipage}{0.49\textwidth}
  %   \centering
  %   \includegraphics[width=\linewidth]{content/figures/soft_f1.png}
  %   \caption{Soft-F1 Score}
  %   \label{fig:soft_f1}
  % \end{minipage}
  \begin{minipage}{0.49\textwidth}
    \centering
    \includegraphics[width=\linewidth]{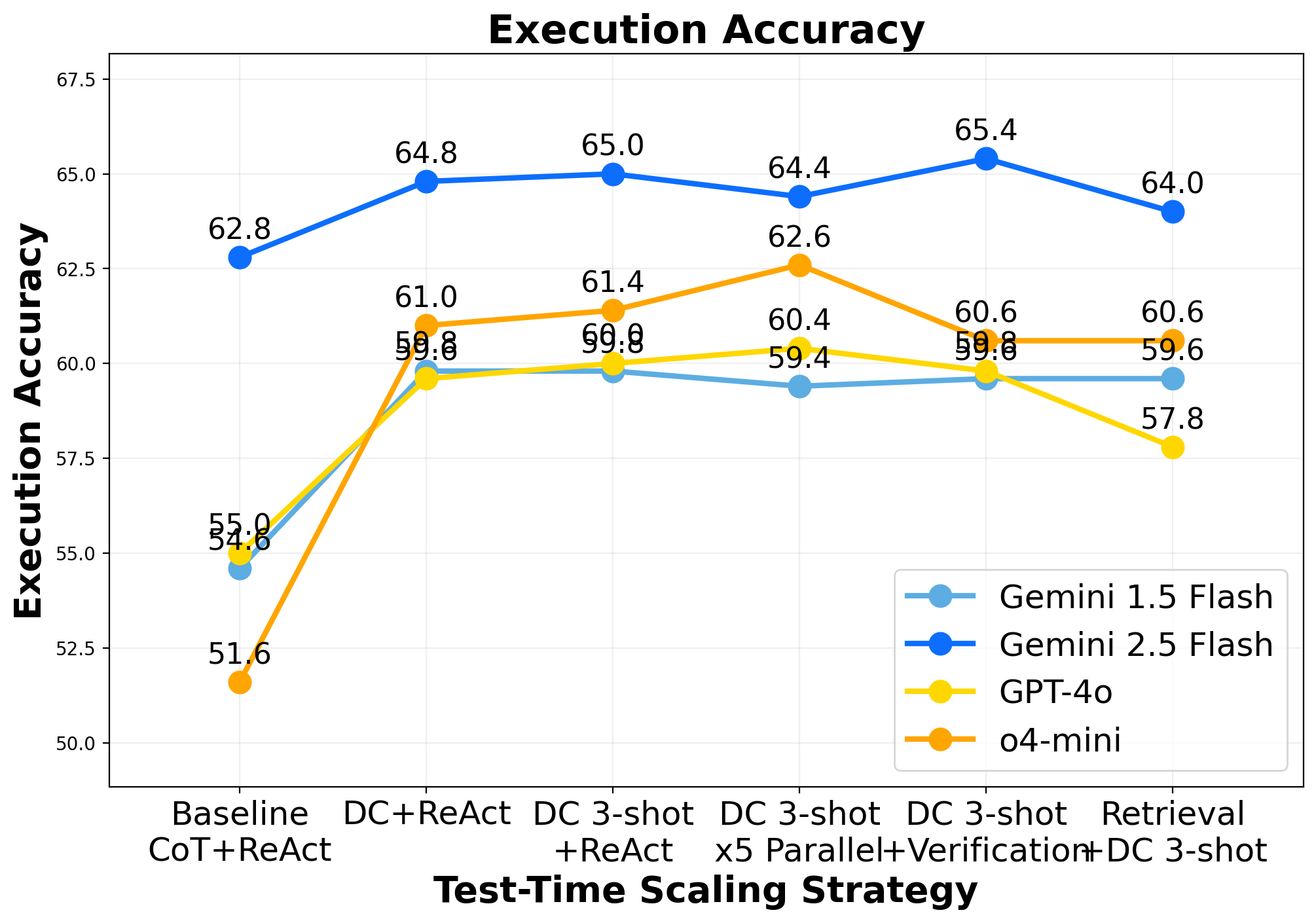}
  \end{minipage}
  \hfill
  \begin{minipage}{0.49\textwidth}
    \centering
    \includegraphics[width=\linewidth]{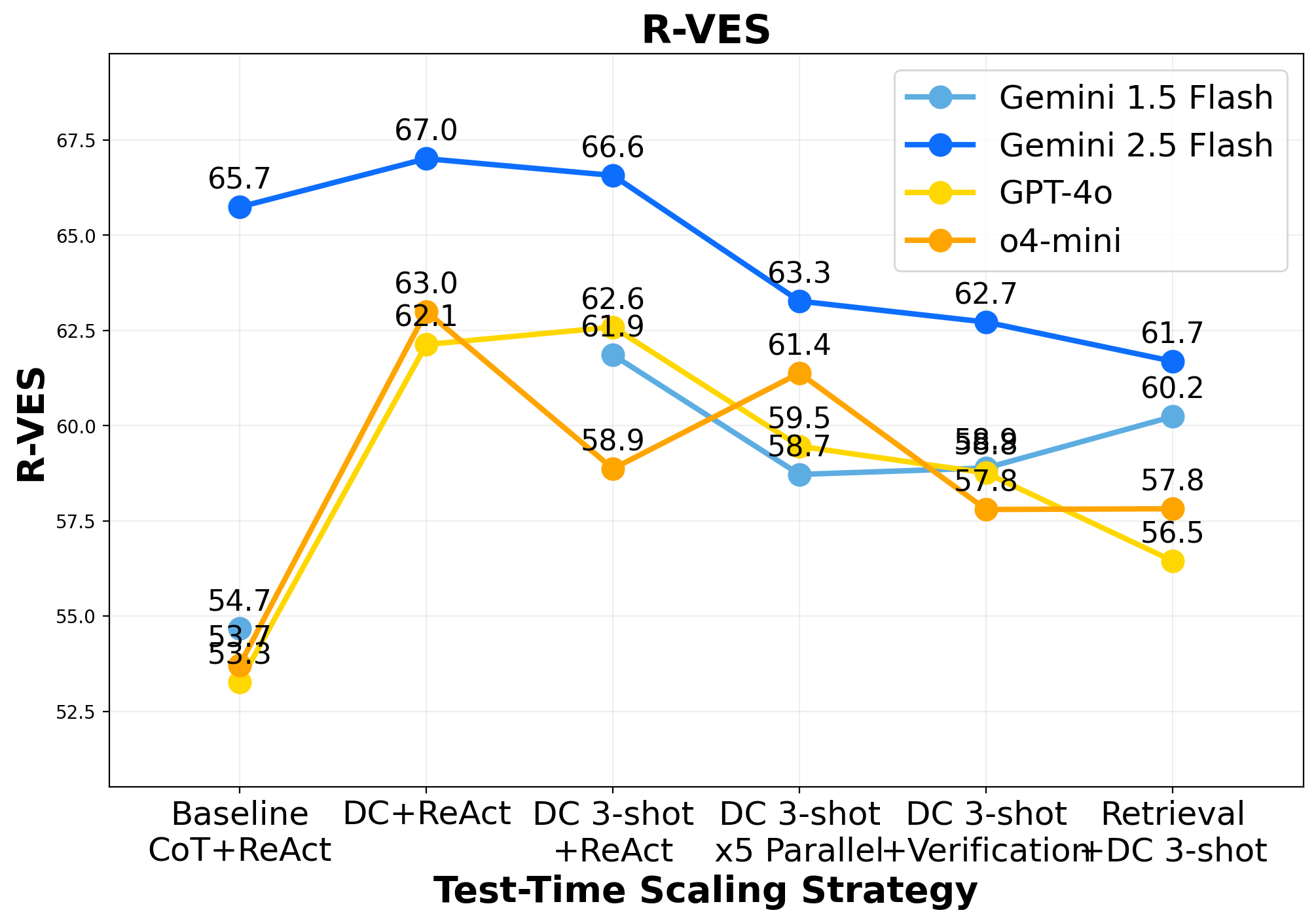}
  \end{minipage}
\caption{Execution Accuracy and R-VES Score}
\label{fig:ex_r_ves}
\end{figure}

The results of EX and R-VES are visualized in Figure~\ref{fig:ex_r_ves}.

\begin{figure}[t]
    \centering
    \includegraphics[width=\linewidth]{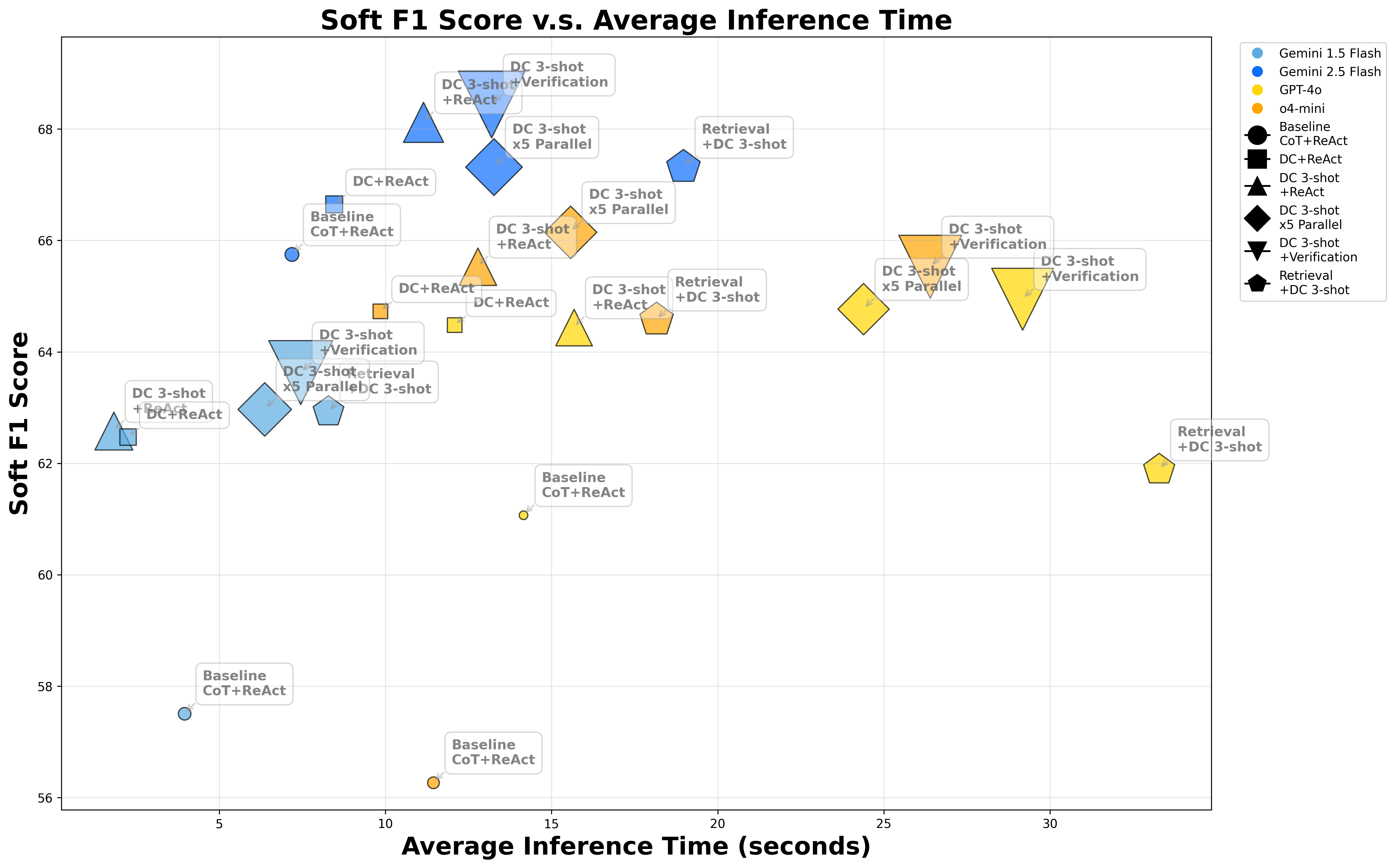}
    \caption{Soft-F1 Score by Average Inference Time}
    \label{fig:soft_f1_inf}
\end{figure}

\begin{figure}[t]
    \centering
    \includegraphics[width=\linewidth]{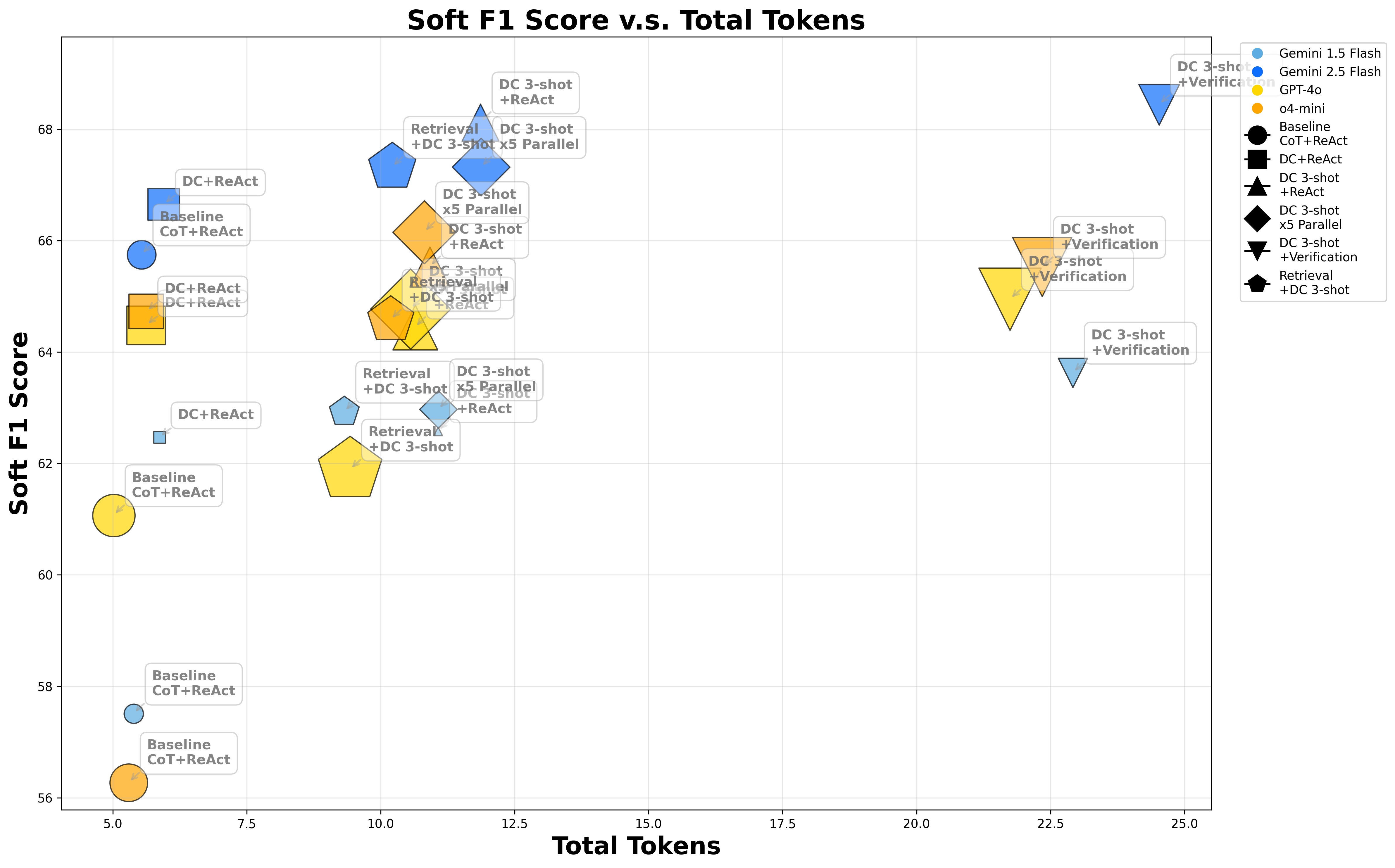}
    \caption{Soft-F1 Score by Average Token Usage (Prompt and Completion). 
    \texttt{DC 3-shot x5 Parallel} reports the token usage of final selected trace.
    }
    \label{fig:soft_f1_token}
\end{figure}

Detailed experiment results are reported in Table~\ref{tab:experiment_result}.
The accuracy and performance results are visualized in Figure~\ref{fig:soft_f1_inf} and Figure~\ref{fig:soft_f1_token}
\input{content/table.exp.01.results}

% \begin{figure}[htbp]
%   \centering
%   \begin{minipage}{0.49\textwidth}
%     \centering
%     \includegraphics[width=\linewidth]{content/figures/execution_error_count.png}
%     \caption{Execution Error Count}
%     \label{fig:ex_inf}
%   \end{minipage}
%   \hfill
%   \begin{minipage}{0.49\textwidth}
%     \centering
%     \includegraphics[width=\linewidth]{content/figures/average_#_of_llm_calls.png}
%     \caption{Average LLM Calls Count.}
%     \label{fig:num_llm_call}
%   \end{minipage}
% \end{figure}

%% file: content/table.exp.01.results.tex
\begin{table}[h!]
\centering
\small
\begin{tabular}{llllllll}
\toprule
\textbf{Workflow} & \textbf{Model} & \textbf{EX} & \textbf{Soft-F1} & \textbf{R-VES} & \textbf{Inf (s)} & \textbf{PT (k)} & \textbf{CT (k)} \\
\midrule
Baseline CoT+ReAct    & Gemini 1.5 Flash & 54.6 & 57.51 & 54.68 &  3.95 &  5.36 & 0.08 \\
Baseline CoT+ReAct    & Gemini 2.5 Flash & 62.8 & 65.75 & 65.74 &  7.19 &  5.18 & 0.26 \\
Baseline CoT+ReAct    & GPT-4o           & 55.0 & 61.07 & 53.27 & 14.15 &  4.61 & 0.57 \\
Baseline CoT+ReAct    & o4-mini          & 51.6 & 56.27 & 53.72 & 11.45 &  4.60 & 0.74 \\

DC+ReAct               & Gemini 1.5 Flash & 59.8 & 62.47 & 57.18 &  2.25 &  5.80 & 0.08 \\
DC+ReAct               & Gemini 2.5 Flash & 64.8 & 66.65 & 67.01 & 10.95 &  4.39 & 0.15 \\
DC+ReAct               & GPT-4o           & 59.6 & 64.48 & 62.13 & 14.77 &  4.96 & 0.58 \\
DC+ReAct               & o4-mini          & 61.0 & 64.73 & 63.00 & 12.65 &  4.47 & 1.02 \\

DC 3-shot+ReAct      & Gemini 1.5 Flash & 59.8 & 62.58 & 61.86 &  4.12 &  9.72 & 0.15 \\
DC 3-shot+ReAct      & Gemini 2.5 Flash & 65.0 & 68.12 & 64.26 & 14.61 &  8.42 & 0.19 \\
DC 3-shot+ReAct      & GPT-4o           & 60.0 & 64.44 & 63.07 & 18.08 &  6.71 & 0.50 \\
DC 3-shot+ReAct      & o4-mini          & 61.4 & 65.53 & 62.02 & 17.12 &  5.02 & 1.35 \\

DC 3-shotx5 Parallel & Gemini 1.5 Flash & 59.4 & 63.31 & 59.76 &  5.93 & 10.97 & 0.12 \\
DC 3-shotx5 Parallel & Gemini 2.5 Flash & 63.0 & 66.00 & 63.06 & 17.24 &  9.53 & 0.12 \\
DC 3-shotx5 Parallel & GPT-4o           & 59.2 & 64.91 & 57.73 & 17.94 &  7.94 & 0.62 \\
DC 3-shotx5 Parallel & o4-mini          &  62.6  &   66.15  &   61.38   &  15.57   &   9.87  &  1.08   \\

DC 3-shot+Verification & Gemini 1.5 Flash & 59.2 & 63.24 & 58.27 & 10.92 & 12.23 & 0.34 \\
DC 3-shot+Verification & Gemini 2.5 Flash & 64.8 & 68.28 & 65.24 & 20.02 & 10.39 & 1.09 \\
DC 3-shot+Verification & GPT-4o           & 56.6 & 60.68 & 59.42 & 28.16 &  8.47 & 0.86 \\
DC 3-shot+Verification & o4-mini          & 59.4 & 63.56 & 57.00 & 26.68 &  7.85 & 1.98 \\

Retrieval+DC 3-shot  & Gemini 1.5 Flash & 59.6 & 62.93 & 60.24 &  8.28 &  9.22 & 0.11 \\
Retrieval+DC 3-shot  & Gemini 2.5 Flash & 64.0 & 67.32 & 61.69 & 18.97 &  9.39 & 0.76 \\
Retrieval+DC 3-shot  & GPT-4o           & 57.8 & 62.74 & 57.56 & \textbf{33.28} &  8.71 & 0.72 \\
Retrieval+DC 3-shot  & o4-mini          & 60.6 & 64.58 & 57.82 & 18.16 &  8.61 & \textbf{1.67} \\
\bottomrule
\end{tabular}
\caption{Experiment Results}
\label{tab:experiment_result}
\end{table}